\documentclass[runningheads]{llncs}
\usepackage[T1]{fontenc}
\usepackage{graphicx}
\usepackage{calc}
\usepackage{algorithm}
\usepackage{amsmath}
\usepackage{subfigure}
\usepackage[noend]{algpseudocode}
\usepackage{float}
\usepackage{stackengine}
\usepackage{multirow}
\usepackage{booktabs}
\usepackage{tikz}
\usepackage{amssymb}
\usepackage{tabularx}
\usepackage{arydshln}
\usepackage{siunitx}

\usepackage{xcolor}

\usepackage[hidelinks,colorlinks=true,linkcolor=blue,citecolor=blue]{hyperref}
\usepackage{makecell} 

\usepackage[capitalize]{cleveref}
\crefname{section}{Sec.}{Secs.}
\Crefname{section}{Section}{Sections}
\Crefname{table}{Table}{Tables}
\crefname{table}{Tab.}{Tabs.}

\makeatletter
\renewcommand{\@fnsymbol}[1]{%
  \ifcase#1 \or *\or \dag\or \ddag\or \S\or \P\or \|\or **\or \dag\dag \or \ddag\ddag \fi}
\makeatother

\institute{
Department of Electronic and Computer Engineering, The Hong Kong University of Science and Technology, Hong Kong SAR, China \and
Division of Paediatric Dentistry and Orthodontics, Faculty of Dentistry, The University of Hong Kong, Hong Kong SAR, China
}

\begin{document}
\title{Geometric-Guided Few-Shot Dental Landmark Detection with Human-Centric Foundation Model}
\titlerunning{GeoSapiens}
%

\newcommand{\repeatthanks}{\textsuperscript{\thefootnote}}

\author{
Anbang Wang\inst{1}\thanks{Equal contribution.} 
\and
Marawan Elbatel\inst{1}\repeatthanks \and
Keyuan Liu\inst{2} \and
Lizhuo Lin\inst{2} \and
Meng Lan\inst{1} \and
Yanqi Yang\inst{2}\thanks{Corresponding authors: \email{eexmli@ust.hk}, \email{yangyanq@hku.hk}} \and
Xiaomeng Li\inst{1}\repeatthanks
}
\authorrunning{A. Wang et al.}
\maketitle              

\begin{abstract}
Accurate detection of anatomic landmarks is essential for assessing alveolar bone and root conditions, thereby optimizing clinical outcomes in orthodontics, periodontics, and implant dentistry. Manual annotation of landmarks on cone-beam computed tomography (CBCT) by dentists is time-consuming, labor-intensive, and subject to inter-observer variability. Deep learning-based automated methods present a promising approach to streamline this process efficiently. However, the scarcity of training data and the high cost of expert annotations hinder the adoption of conventional deep learning techniques. To overcome these challenges, we introduce GeoSapiens, a novel few-shot learning framework designed for robust dental landmark detection using limited annotated CBCT of anterior teeth. Our GeoSapiens framework comprises two key components: (1) a robust baseline adapted from Sapiens, a foundational model that has achieved state-of-the-art performance in human-centric vision tasks, and (2) a novel geometric loss function that improves the model's capacity to capture critical geometric relationships among anatomical structures. Experiments conducted on our collected dataset of anterior teeth landmarks revealed that GeoSapiens surpassed existing landmark detection methods, outperforming the leading approach by an 8.18\% higher success detection rate at a strict 0.5 mm threshold-a standard widely recognized in dental diagnostics. Code is available at: \url{https://github.com/xmed-lab/GeoSapiens}.

\end{abstract}
\keywords{Few-shot Learning  \and Foundation Model 
\and Landmark Detection}

%
%
%
\section{Introduction}

Evaluating the condition of alveolar bone and tooth roots is a key aspect in orthodontics, periodontics, and implant dentistry. Precise landmark detection, including the cementoenamel junction, physiological crest and root apex, is crucial for measuring clinical metrics, as even a 0.5 mm error in identifying the cementoenamel junction (CEJ) or alveolar crest could lead to misclassification-either falsely diagnosing dehiscence (CEJ-to-Bone Crest >2 mm) or failing to detect an early-stage defect, directly impacting treatment decisions~\cite{teeth_1,teeth_2}. However, comprehensive measurement of the clinical metrics for individual teeth presents significant challenges. This process of identifying critical anatomical landmarks on CBCT images is time-consuming, labor-intensive, and prone to variability. Recently, deep learning techniques have demonstrated remarkable success in pose-estimation~\cite{deeppose} and landmark detection tasks~\cite{landmark}, showing significant potential for automating our dental landmark detection process. However, the scarcity of annotated datasets poses a significant barrier to deep learning implementation, as the manual annotation process for these specialized landmarks requires substantial expert time and resources. Therefore, developing effective few-shot learning approaches for dental landmark detection becomes extremely crucial.

There have been various deep learning-based works utilizing dental CBCT images, in order to detect and classify~\cite{detection_class} or segment~\cite{segmentation} teeth structures. Several studies have also employed intraoral tooth images to detect bone dehiscence and fenestration, with ground-truth labels derived from CBCT imaging~\cite{fdsos,Liu2025FDTooth}. However, there is no prior work focusing on landmarks detection of anterior teeth CBCT images. To this end, we collected data from patients who underwent orthodontic treatment and created the LDTeeth dataset. For each image, we obtained the bounding box information of the correct teeth and corresponding annotations of 16 landmarks on the dataset. The detailed annotation process will be discussed in~\ref{Section:dataset}.

A straightforward approach to dental landmark detection would be applying conventional landmark detection techniques. However, these traditional methods generally demand substantial training datasets, and they exhibit considerable performance deterioration when implemented in limited-data scenarios. To deal with this issue, FM-OSD~\cite{Mia_FMOSD_MICCAI2024} applied template-based matching methods. While being effective on other datasets, FM-OSD struggles with dental landmark detection due to the complexity and variability of features in this task. Another approach, GU2Net~\cite{GU2Net}, involves leveraging cross-dataset information for training. Nevertheless, all current approaches have shown limited success in few-shot dental landmark detection. (see~\cref{tbl:sota_table})

Recent advancements in \textbf{vision foundation models} have demonstrated state-of-the-art (SOTA) performance across various visual tasks~\cite{mllm}, particularly excelling in few-shot scenarios with remarkable generalization capabilities~\cite{SAM,DINO}. However, applying these models to dental landmarks detection faces significant challenges: the scarcity of anterior tooth CBCT images and the lack of foundation models tailored for this task make it difficult to select an appropriate baseline.

To address these limitations, we propose GeoSapiens, a novel few-shot dental landmarks detection framework built upon Sapiens, a foundation vision model that achieves SOTA performance in human-centric tasks. 
\begin{table}[h]
  \centering
  \caption{Information about our collected LDTeeth dataset.}
  \label{tbl:ldteeth_dataset}
  \resizebox{\textwidth}{!}{
    \begin{tabular}{c|c|c|c|c|c|c|c|c|c|c}
      \hline
      \multirow{3}{*}{\begin{tabular}[c]{@{}c@{}}Number of\\images\end{tabular}} & 
      \multirow{3}{*}{Resolution} & 
      \multirow{3}{*}{\begin{tabular}[c]{@{}c@{}}Bbox\\Annotation\end{tabular}} & 
      \multirow{3}{*}{\begin{tabular}[c]{@{}c@{}}Landmark\\Annotation\end{tabular}} & 
      \multirow{3}{*}{\begin{tabular}[c]{@{}c@{}}Number of\\Landmarks\end{tabular}} & 
      \multicolumn{6}{c}{Split} \\
      \cline{6-11}
      & & & & & \multicolumn{2}{c|}{Train} & \multicolumn{2}{c|}{Val} & \multicolumn{2}{c}{Test} \\
      \cline{6-11}
      & & & & & Images & Patients & Images & Patients & Images & Patients \\
      \hline
      347 & 957$\times$555 & \checkmark & \checkmark & 5,552 (16 per image) & 36 & 3 & 149 & 14 & 162 & 16 \\
      \hline
    \end{tabular}
    
  }
\end{table}
\begin{figure}[h]
    \centering
    \vspace{-0.3cm}
    \includegraphics[width=\textwidth]{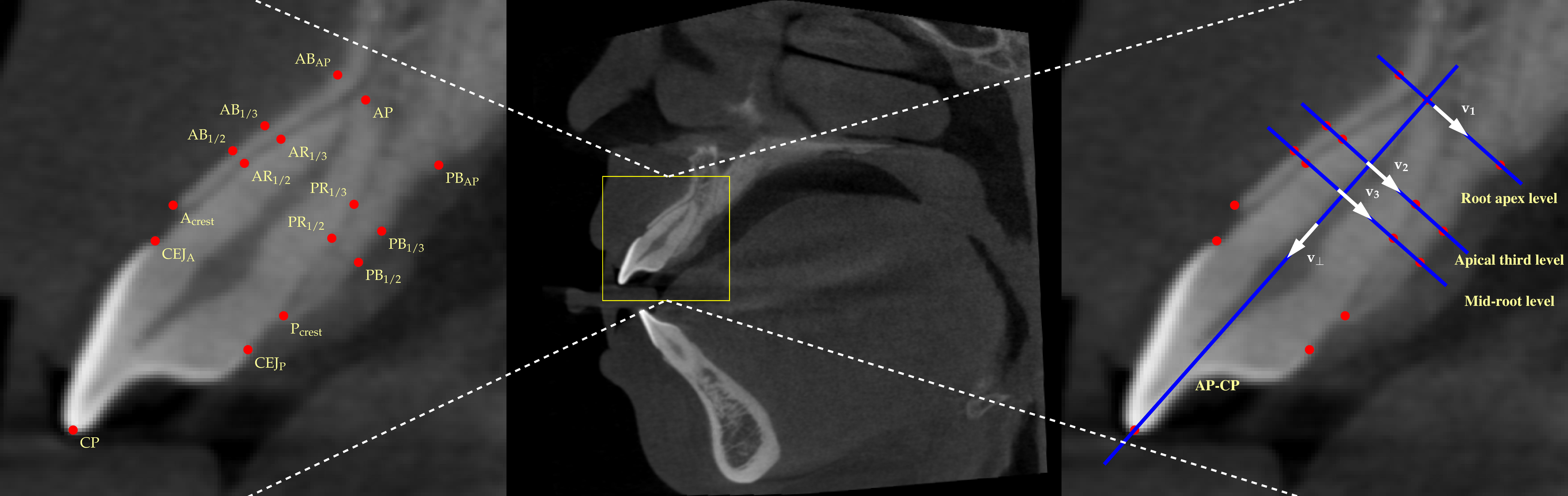}
    \caption{The image in our dataset contains both upper teeth and lower teeth (middle). The teeth containing a complete mid-sagittal plane section was labeled by a bounding box and corresponding landmarks were annotated (left). AP-CP, Root apex level, Apical third level, and Mid-root level lines are shown by blue lines (right). The geometric loss encourages the model to predict lines following proper geometric features by measuring dot product between each unit vectors. }
    \label{fig:dataset_sample}
    \vspace{-0.6cm}
\end{figure}
Our motivation to introduce this foundation model to our task can be attributed to its compelling performance in pose-estimation task and the similarity between the symmetric feature appearing in both teeth and the human body. To make training practical, we incorporate LoRA (Low-Rank Adaptation), a parameter-efficient fine-tuning (PEFT) method, reducing the trainable parameters of the ViT backbone to just 0.3\% of the original count~\cite{LoRA}. Furthermore, we recognize that landmarks detection inherently relies on geometric relationships-such as the spatial correlations among anterior teeth landmarks-which are underexploited in generic models. Motivated by this, we introduce a novel geometric loss function to explicitly encode these priors, enhancing the models' ability to capture task-specific characteristics.

In summary, our proposed work, \textbf{GeoSapiens}, makes the following key contributions:
(1) We established the first anterior teeth CBCT dataset specifically designed for dental landmark detection tasks. (2) We discovered limitations in current few-shot methods and devised a strong baseline utilizing a foundation model, Sapiens. (3) We propose a novel geometric-based loss function, enhancing model generalization and robustness by incorporating structural priors, leading to improved accuracy in challenging scenarios. Through extensive experiments, our method achieved the best performance among all other methods, surpassing the current best-performing method by 8.18\% success detection rate under the 0.5 mm threshold.

\section{Method}

\subsection{LDTeeth Dataset}
\label{Section:dataset}

During the dataset collection, for each patient, a lateral CBCT scan of the dental region was performed, and from the anterior tooth images, doctor selected slices with a correct mid-sagittal plane section of the teeth accurately and efficiently using CBCT viewing software, consistent with previous work~\cite{Sun2022}. These selected slices, which contain a complete mid-sagittal plane section, as depicted in~\cref{fig:dataset_sample}, were then annotated. 
During the annotation, dentists initially annotated six points for each anterior tooth image, marking the crown point (CP), apical point (AP), the labial CEJ point (CEJ$_{\text{A}}$), lingual CEJ point (CEJ$_{\text{P}}$) and the labial and lingual points of the alveolar crest (A$_\text{crest}$ and P$_\text{crest}$). The machine then generated perpendicular lines to the tooth axis (CP-AP) through the AP point, and then at levels 1/2 and 1/3 of the root length. Subsequently, the machine automatically marked the intersection points of these lines with the labial (A) and lingual (P) alveolar bone (B) and root (R), totaling 10 additional points: AB$_{\text{AP}}$, AB$_{1/3}$, AR$_{1/3}$, AB$_{1/2}$, AR$_{1/2}$, PB$_{\text{AP}}$, PB$_{1/3}$, PR$_{1/3}$, PB$_{1/2}$, PR$_{1/2}$. Afterward, an expert manually corrected the points. Obtaining a total $16$ points for an image.

It is important to note that our dataset includes both upper and lower teeth. Additionally, the anterior teeth exhibit significant morphological differences based on their position within the dental arch. These variations lead to differences in image features, making it highly challenging for the model to effectively capture these features in a few-shot setting.

\begin{figure}[t]
    \centering
    \includegraphics[width=\textwidth]{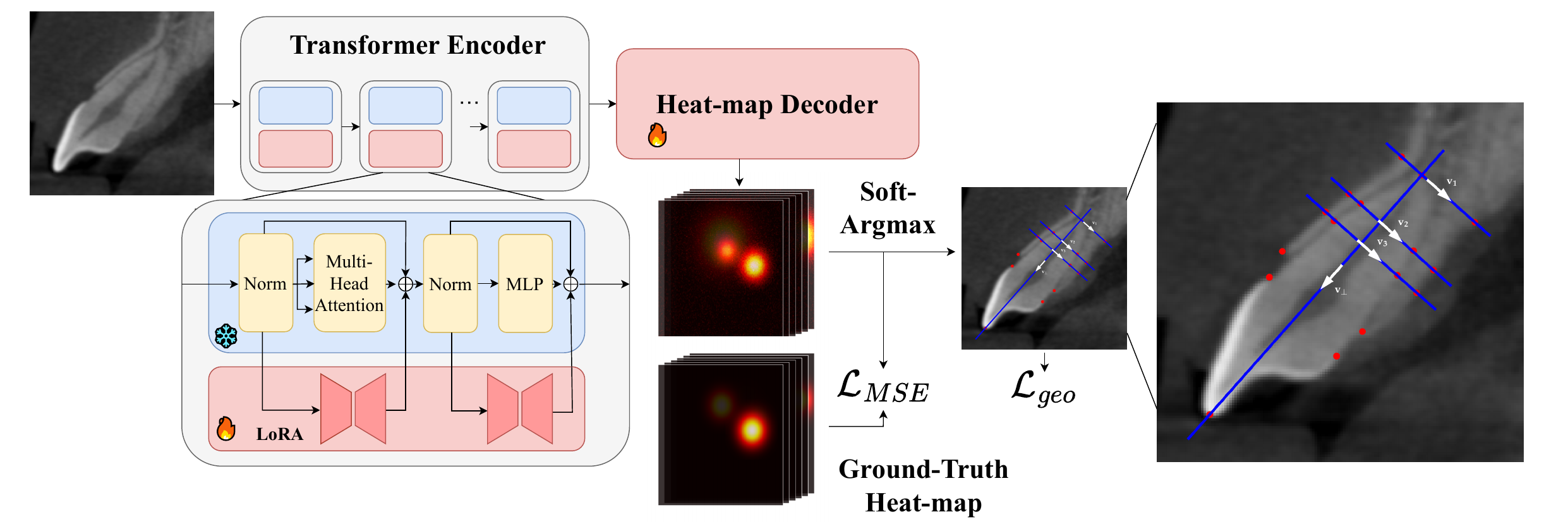}
    \caption{Overall framework of our GeoSapiens.}
    \label{fig:overall_framework}
\end{figure}

\subsection{Overall framework} 
As illustrated in Figure~\ref{fig:overall_framework}, our approach begins with feature extraction from CBCT images using a Transformer encoder. This Vision Transformer (ViT) backbone leverages pre-trained weights from Sapiens, a foundation model renowned for its excellence in human-centric vision tasks. Subsequently, a heatmap decoder predicts heatmaps from these features, followed by the application of Soft-argmax to generate probabilistic landmark coordinates. These coordinates inform our novel geometric loss, which, when combined with mean squared error (MSE) loss, optimizes the model. To enhance efficiency, we employ Low-Rank Adaptation (LoRA) for fine-tuning, substantially reducing the number of trainable parameters and the associated training cost.

\textbf{Sapiens} is a family of models designed for human-centric vision tasks~\cite{Sapiens}, pre-trained using a Masked Autoencoder (MAE) on the Humans-300M dataset of 300 million in-the-wild human images. This self-supervised approach masks portions of images and trains the model to reconstruct them, yielding robust visual feature representations effective across various tasks. The exceptional performance of the Sapiens model on our dataset can potentially be attributed to the similarities in feature distributions between its pre-trained dataset and our target dataset. Our CBCT images of anterior teeth exhibit inherent symmetrical characteristics that are consistent with the structural patterns commonly found in human-centric image datasets. These shared symmetrical and morphological features facilitate a more stable fine-tuning process and enhance the model's generalization capability. 

\subsection{Geometric loss}

Although the existing framework yields impressive results, we observed that our model struggles to capture the geometric features of the dataset, resulting in a suboptimal solution. To address this, we developed a geometric loss function to enhance the model's ability to discern these features. Specifically, guided by the data annotation method, the model is trained to recognize the perpendicular relationship between the AP-CP line and three horizontal lines-Root Apex Level, Apical Third Level, and Mid-root Level-while ensuring parallelism among these three lines, as illustrated in Figure~\ref{fig:dataset_sample}.

In heat-map-based landmark detection tasks, a discrepancy exists between the processing methods employed during training and inference phases. During inference, landmark coordinates are typically extracted from the heat-map using the arg-max operation. However, as arg-max is non-differentiable and unsuitable for gradient propagation, it cannot be applied directly during training. Inspired by IntegralNet~\cite{IntegralNet}, we adopt the soft-argmax operation to extract coordinates differentiably during training, enabling the integration of coordinate information and facilitating the subsequent incorporation of geometric loss functions.

More specifically, we treat the keypoint coordinates as the expected value of the coordinates over the heat-map:

\begin{equation}
\hat{p} = \sum_{p} p \cdot M(p)
\end{equation}

where $ p $ represents the coordinates of a point, and $ M(p) $ denotes the normalized probability at that point. The probabilities are obtained using the softmax function:
\begin{equation}
    M(p) = \frac{e^{\frac{H(p)}{T}}}{\sum_{q} e^{\frac{H(q)}{T}}}
\end{equation}
where $ H(p) $ is the heat-map value at point $ p $. $T$ is the temperature. The smaller value of $T$ leads the predicted coordinate similar to the location of arg-max, while this will exploit less information with neighboring probability. Larger values overly smooth predictions, collapsing coordinates toward the center. In our experiment, we set $T=0.1$, which was effective with minimal sensitivity nearby. 

\noindent{\textbf{Loss function}
After obtaining the point coordinates, we obtain the lines using linear regression, and obtain the unit direction vector of each line. We denote the unit vectors of line AP-CP, lines of Root apex level, Apical third level, and Mid-root level lines as \(\mathbf{v_{\perp}}, \mathbf{v_{1}}, \mathbf{v_{2}},\mathbf{v_{3}}\). The loss function is defined as
\begin{equation}
    \mathcal{L}_{\text{geo}} = \frac{1}{6}(\sum_{j=1}^{3} \mathbf{v_{\perp}} \cdot \mathbf{v_{j}} + \sum_{j=1}^{2}\sum_{k=j+1}^{3}(1-|\mathbf{v_{j} \cdot \mathbf{v}_{k}}|))
\end{equation}
The geometric loss is designed to enforce two types of geometric constraints: perpendicularity and parallelism. For perpendicularity, we use the dot product $\mathbf{v_{\perp}} \cdot \mathbf{v_{j}}$ which equals zero when the vectors are orthogonal. For parallelism, we use $1 - |\mathbf{v_{j}} \cdot \mathbf{v_{k}}|$, which penalizes deviations from perfect alignment (where the dot product approaches 1 or -1). The normalization factor $\frac{1}{6}$ ensures that each term contributes equally to the total loss.

\section{Experiment}

\noindent{\textbf{Comparative Methods.}}
We compare our GeoSapiens with SOTA models in few-shot anatomic landmark detections, including GU2Net~\cite{GU2Net}, FM-OSD~\cite{Mia_FMOSD_MICCAI2024}, NFDF~\cite{TMI24_generative_prior_landmark}. For FM-OSD, we adapt the one-shot model to a few-shot by adding more images to the training data. Since both upper teeth and lower teeth appears in our dataset, we use two templates from upper teeth and lower teeth respectively to do comparison. For GU2Net, we add our LDTeeth dataset to original three datasets in addition to head, hand, and chest.\\
\noindent{\textbf{Implementation details.}}
To represent the few-shot setting, we utilized a training set containing images from only 3 patients. For GeoSapiens, we utilize Sapiens-0.3B~\cite{Sapiens} and initialize it with pre-trained weights. Training is performed using the AdamW optimizer with a learning rate of $5 \times 10^{-4}$. The learning rate schedule comprises two phases: an initial linear warm-up over the first 500 steps, scaling from a factor of 0.001 to the base rate, followed by a multi-step decay with reductions by a factor of 0.1 at epochs 170 and 200. A batch size of 16 is maintained throughout training. For LoRA fine-tuning, we configure query-key-value projections in attention layers with $\alpha = 4$ and $r = 4$, and projection layers with $\alpha = 8$ and $r = 8$. The total loss is defined as $\mathcal{L}{_\text{total}} = \mathcal{L}{_\text{MSE}} + \lambda \mathcal{L}_{\text{geo}}$. Since the MSE loss in our heatmap-based regression is computed over all pixels in the output heatmap, where the vast majority correspond to easily identifiable background regions (i.e., ground truth value 0), the averaged MSE becomes naturally small. In contrast, our geometric loss imposes a global structural constraint across landmarks. To balance their magnitudes and ensure stable gradient flow during optimization, we assign a small weight to geometric loss, with $\lambda=10^{-5}$. For the heat-map decoder, we adopt the top-down heat-map head proposed in~\cite{simplebaseline}.\\
\noindent{\textbf{Evaluation metrics}}
The model's performance is evaluated using two primary metrics: mean radial error (MRE) and successful detection rate (SDR). The MRE measures the average Euclidean distance between predicted and ground-truth landmark coordinates, while the SDR quantifies the proportion of predictions falling within specified distance thresholds. To assess performance under strict accuracy requirements, we set thresholds at 0.5 mm, 1.0 mm, and 2.0 mm in our experiments, with the 0.5 mm threshold emphasized due to its widespread adoption in clinical practice and value for dentistry~\cite{teeth_1}. For a concise summary of the SDR, we introduce SDR$_{\text{average}}$, defined as the mean of the SDR values across these three thresholds.



\begin{table*}[t]
    \centering
    \caption{Comparision with SOTA methods on our LDTeeth dataset.}
    \label{tab:combined_results}
    \begin{tabular}{@{}l *{5}{c}@{}} 
    \toprule
    \multirow{2}{*}{Method} & 
    \multicolumn{3}{c}{SDR (\unit{\%})} &
    \multicolumn{1}{c}{\multirow{2}{*}{SDR$_{\text{average}}$ (\unit{\%})}} &
    \multicolumn{1}{c}{\multirow{2}{*}{MRE (\unit{mm})}} \\
    \cmidrule(lr){2-4}
    & 0.5mm & 1mm & 2mm & & \\
    \midrule
    GU2Net~\cite{GU2Net} & 45.21 & 64.12 & 81.90 & 63.74 & 1.312 \\
    FM-OSD~\cite{Mia_FMOSD_MICCAI2024} & 32.95 & 55.79 & 77.62 & 55.45 & 1.520 \\
    NFDP~\cite{TMI24_generative_prior_landmark} & 55.01& 80.40 & 92.09 & 75.83 & 0.825 \\

    \textbf{GeoSapiens (ours)} & \textbf{63.19} & \textbf{84.14} & \textbf{93.36} & \textbf{80.23} & \textbf{0.747} \\
    \bottomrule
    \end{tabular}
    \label{tbl:sota_table}
\end{table*}

\subsection{Performance on Few-shot landmarks detection}

Table~\ref{tbl:sota_table} presents a performance comparison between our method and other state-of-the-art approaches on the LDTeeth dataset. GU2Net exhibits underwhelming performance, with an SDR$_{\text{average}}$ of 63.74\% and an MRE of 1.312 mm. Despite training on multiple datasets, its effectiveness is constrained by insufficient domain-specific data. FM-OSD performs even worse, achieving an SDR$_{\text{average}}$ of 55.45\% and an MRE of 1.520 mm, primarily due to its template's inability to capture the complex features of our dataset, which hinders accurate matching. Although NFDP achieves a relatively high SDR of 92.09\% at the 2.0 mm threshold and an MRE of 0.825 mm, it struggles with high-precision detection, recording SDRs of only 55.01\% at 0.5 mm and 80.40\% at 1.0 mm, rendering it impractical for clinical applications.

\textbf{GeoSapiens} establishes new state-of-the-art benchmarks across all metrics, with SDRs of 63.19\% at 0.5 mm, 84.14\% at 1.0 mm, and 93.36\% at 2.0 mm, an SDR$_{\text{average}}$ of 80.23\%, and an MRE of 0.747 mm. It surpasses the widely used cross-dataset method GU2Net, improving SDR$_{\text{average}}$ by \textbf{16.49\%} and reducing MRE by \textbf{0.565 mm}. Compared to the leading template-matching method FM-OSD, GeoSapiens increases SDR$_{\text{average}}$ by \textbf{24.78\%} and decreases MRE by \textbf{0.773 mm}. Against NFDP, GeoSapiens improves SDR by \textbf{8.18\%} at 0.5 mm and \textbf{3.74\%} at 1.0 mm, while lowering MRE by \textbf{0.078 mm}, enhancing its suitability for clinical use. These results underscore GeoSapiens’ superior performance in few-shot dental landmark detection. Qualitative results and the geometric loss curve are depicted in~\cref{fig:qualitative}.

\noindent \textbf{Ablation Study.}
To assess the effectiveness of our proposed components, we conducted comprehensive ablation studies on our framework, as presented in Table~\ref{tbl:ablation}. Our baseline model, Sapiens, exhibits strong performance, achieving SDRs of 62.77\% at 0.5 mm, 84.29\% at 1.0 mm, and 93.75\% at 2.0 mm, with an SDR$_{\text{average}}$ of 80.27\% and an MRE of 0.740 mm. The addition of our geometric loss function further improves performance, yielding increases of \textbf{2.50\%} and \textbf{1.86\%} at the 0.5 mm and 1.0 mm thresholds, respectively. However, fully fine-tuning the model with these enhancements requires 330 million trainable parameters, rendering it impractical for most clinical settings. To mitigate this, we integrate LoRA, reducing the trainable parameters from 330 million to 24 million-a \textbf{92.73\%} reduction. Although applying LoRA alone slightly degrades performance, the inclusion of our geometric loss function robustly enhances the model. Specifically, 

GeoSapiens achieves SDRs of 63.19\% at 0.5 mm (a \textbf{0.39\%} increase) and 84.14\% at 1.0 mm (a \textbf{0.85\%} increase from the Baseline-LoRA) and reduces the MRE by \textbf{1.96\%}, demonstrating consistent and robust improvements.

\begin{table}[t]
\centering
\caption{Ablation study on our proposed framework.}
\resizebox{\columnwidth}{!}{
\begin{tabular}{c|c|c|c|ccc|c|c}
\toprule
\multirow{2}{*}{Method} & \multirow{2}{*}{LoRA} & \multirow{2}{*}{$\mathcal{L}_{geo}$} & \# Trainable & \multicolumn{3}{c|}{SDR (\%)} & \multirow{2}{*}{SDR$_{\text{average}}$ (\%)} & \multirow{2}{*}{MRE (\unit{mm})} \\
\cline{5-7}  
 & & & Parameters& 0.5\unit{mm} & 1\unit{mm} & 2\unit{mm} & & \\
\hline
Baseline  & $\times$ & $\times$ & 330M & 62.77 & 84.29 & 93.75 & 80.27 & 0.740 \\
Baseline-Geometric  & $\times$ & $\checkmark$ & 330M & 65.27 & 86.15 & 93.59 & 81.67 & 0.742 \\
\hline
Baseline-LoRA & \checkmark & $\times$ & 24M & 62.80 & 83.29 & 93.24 & 79.78 & 0.762 \\
GeoSapiens & \checkmark & \checkmark & 24M & 63.19 & 84.14 & 93.36 & 80.23 & 0.747 \\
\bottomrule
\end{tabular}}
\label{tbl:ablation}
\end{table}

\begin{figure}[t]
    \centering
    \includegraphics[width=\linewidth]{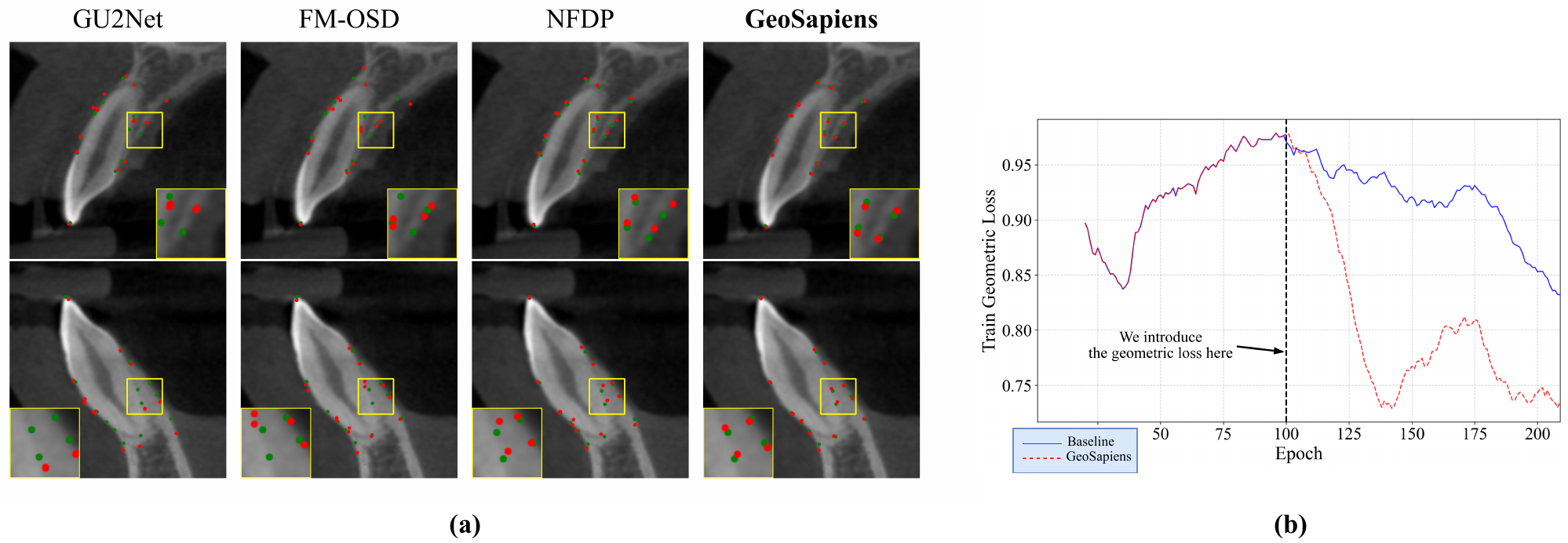}
    \caption{(a) Qualitative results for our \textbf{GeoSapiens} and baselines. Red and green points represent predicted landmarks and ground-truth labels, respectively. We visualize two sample predictions of upper teeth and lower teeth for each method. (b) Geometric loss comparison during our training stage.  }
    \label{fig:qualitative}
\end{figure}

\section{Conclusion}

In this study, we propose an efficient approach for dental landmark detection in CBCT images under few-shot conditions. By leveraging the Sapiens foundational model with LoRA fine-tuning, we achieve state-of-the-art performance while reducing trainable parameters from 330 million to 24 million. The integration of our geometric loss function further enhances the model's capacity to discern anatomical relationships, resulting in superior accuracy, particularly at stringent precision thresholds. Our method substantially outperforms existing techniques in addressing complex dental features, underscoring its potential for practical clinical applications. Comprehensive ablation studies validate the contributions of each component, confirming the robustness and adaptability of our framework.

\begin{credits}
\subsubsection{\ackname} The work presented in this paper was supported by a grant from the Research Grants Council of the Hong Kong Special Administrative Region, China (Project No. R6005-24), as well as a HKUST Undergraduate Research Opportunity Program (UROP) Support Grant.
Anbang Wang also received support from the HKUST UROP Research Travel Sponsorship. Marawan received support from the Hong Kong PhD Fellowship Scheme (HKPFS) provided by the Hong Kong Special Administrative Region, China.

\subsubsection{\discintname}
The authors have no competing interests to declare that are relevant to the content of this article.
\end{credits}
%
%
%
%
\bibliographystyle{splncs04}
\bibliography{main}

\end{document}